\documentclass[twocolumn, switch]{article} 
\pdfoutput=1  

\usepackage{preprint}

\usepackage{amsmath, amsthm, amssymb, amsfonts}

\usepackage[utf8]{inputenc}	
\usepackage[style=ieee, backend=biber]{biblatex}

\usepackage[T1]{fontenc}	
\usepackage{xcolor}		
\usepackage[colorlinks = true,
            linkcolor = purple,
            urlcolor  = blue,
            citecolor = cyan,
            anchorcolor = black]{hyperref}	
\usepackage{booktabs} 		
\usepackage{nicefrac}		
\usepackage{microtype}		
\usepackage{lineno}		
\usepackage{float}			

\usepackage{academicons}

\usepackage{lipsum}		

\usepackage{newfloat}
\DeclareFloatingEnvironment[name={Supplementary Figure}]{suppfigure}
\usepackage{sidecap}
\sidecaptionvpos{figure}{c}

\usepackage{titlesec}
\titlespacing\section{0pt}{12pt plus 3pt minus 3pt}{1pt plus 1pt minus 1pt}
\titlespacing\subsection{0pt}{10pt plus 3pt minus 3pt}{1pt plus 1pt minus 1pt}
\titlespacing\subsubsection{0pt}{8pt plus 3pt minus 3pt}{1pt plus 1pt minus 1pt}

\RequirePackage[nohyperlinks, printonlyused, withpage, nolist]{acronym}
\usepackage{siunitx}
\usepackage{cleveref}
\usepackage{makecell}
\usepackage{xcolor}
\usepackage{subcaption}
\usepackage{tablefootnote}
\usepackage{tikz,xcolor,hyperref}
\usepackage{orcidlink}

\definecolor{lime}{HTML}{A6CE39}
\DeclareRobustCommand{\orcidicon}{
	\begin{tikzpicture}
	\draw[lime, fill=lime] (0,0) 
	circle [radius=0.16] 
	node[white] {{\fontfamily{qag}\selectfont \tiny ID}};
	\draw[white, fill=white] (-0.0625,0.095) 
	circle [radius=0.007];
	\end{tikzpicture}
	\hspace{-2mm}
}
\foreach \x in {A, ..., Z}{\expandafter\xdef\csname orcid\x\endcsname{\noexpand\href{https://orcid.org/\csname orcidauthor\x\endcsname}
			{\noexpand\orcidicon}}
}

\title{Generating peak-aware pseudo-measurements for low-voltage feeders using metadata of distribution system operators}

\usepackage{authblk}

\author[1,2,3]{Manuel Treutlein \orcidlink{0009-0006-1071-341X}}
\author[2]{Marc Schmidt \orcidlink{0000-0002-9127-1708}}
\author[2]{Roman Hahn \orcidlink{0009-0002-2115-3711}}
\author[1]{Matthias Hertel \orcidlink{0000-0002-0814-766X}}
\author[1]{Benedikt Heidrich \orcidlink{0000-0002-1923-0848}}
\author[1]{Ralf Mikut \orcidlink{0000-0001-9100-5496}}
\author[1]{Veit Hagenmeyer \orcidlink{0000-0002-3572-9083}}

\affil[1]{Institute for Automation and Applied Informatics (IAI), Karlsruhe Institute of Technology (KIT)}
\affil[2]{Netze BW GmbH}
\affil[3]{Correspondence address: \href{mailto:manuel.treutlein@partner.kit.edu}{manuel.treutlein@partner.kit.edu}} 

\begin{acronym}
    \acro{DSO}[DSO]{distribution system operator}
    \acro{TSO}[TSO]{transmission system operator}
    \acro{CHP}[CHP]{combined heat and power}
    \acro{ML}[ML]{machine learning}
    \acro{CDF}[CDF]{cumulative distribution function}
    \acro{PDF}[PDF]{probability density function}
    \acro{LR}[LR]{linear regression}
    \acro{MLP}[MLP]{multilayer perceptron}
    \acro{XGBoost}[XGBoost]{eXtreme Gradient Boosting}
    \acro{MAE}[MAE]{mean absolute error}
    \acro{nMAE}[nMAE]{normalized MAE}
    \acro{RMSE}[RMSE]{root mean squared error}
    \acro{nRMSE}[nRMSE]{normalized RMSE}
    \acro{SMAPE}[SMAPE]{symmetric mean absolute percentage error}
    \acro{MAPE}[MAPE]{mean absolute percentage error}
    \acro{R2}[R\textsuperscript{2}]{R\textsuperscript{2}-Score}
    \acro{KDE}[KDE]{kernel density estimation}
    \acro{PMag}[PMag]{Peak Magnitude MAE}
    \acro{PTime}[PTime]{Peak Time MAE}
    \acro{PShape}[PShape]{Peak Shape Error}
    \acro{XAI}[XAI]{Explainable Artificial Intelligence}
    \acro{SHAP}[SHAP]{SHapley Additive exPlanations}
    \acro{PDP}[PDP]{Partial Dependence Plot}
    \acro{DSSE}[DSSE]{Distribution System State Estimation}
    \acro{NILM}[NILM]{Non-Intrusive Load Monitoring}
    \acro{AMI}[AMI]{Advanced Metering Infrastructure}
    \acro{BDEW}[BDEW]{German Association of Energy and Water Industries}
    \acro{TSFM}[TSFM]{Time Series Foundation Model}
    \acro{PV}[PV]{photovoltaic}
    \acro{EV}{electric vehicle}
    \acro{DWD}[DWD]{German Meteorological Service}
    \acro{VAE}[VAE]{Variational Autoencoder}
    \acro{LV}[LV]{low-voltage}
    \acro{MV}[MV]{medium-voltage}
\end{acronym}

\begin{document}

\DeclareSIUnit \kilovoltampere { kVA } 

\twocolumn[ 
\begin{@twocolumnfalse} 

\maketitle


\begin{abstract}
\Acfp{DSO} must cope with new challenges such as the reconstruction of distribution grids along climate neutrality pathways or the ability to manage and control consumption and generation in the grid. In order to meet the challenges, measurements within the distribution grid often form the basis for \acp{DSO}. Hence, it is an urgent problem that measurement devices are not installed in many \acf{LV} grids. In order to overcome this problem, we present an approach to estimate pseudo-measurements for non-measured \ac{LV} feeders based on the metadata of the respective feeder using regression models. The feeder metadata comprise information about the number of grid connection points, the installed power of consumers and producers, and billing data in the downstream \ac{LV} grid. Additionally, we use weather data, calendar data and timestamp information as model features. The existing measurements are used as model target. We extensively evaluate the estimated pseudo-measurements on a large real-world dataset with $2{,}323$ \ac{LV} feeders characterized by both consumption and feed-in. For this purpose, we introduce peak metrics inspired by the BigDEAL challenge for the peak magnitude, timing and shape for both consumption and feed-in. As regression models, we use XGBoost, a \acf{MLP} and a \ac{LR}. We observe that \acs{XGBoost} and \ac{MLP} outperform the \ac{LR}. Furthermore, the results show that the approach adapts to different weather, calendar and timestamp conditions and produces realistic load curves based on the feeder metadata. In the future, the approach can be adapted to other grid levels like substation transformers and can supplement research fields like load modeling, state estimation and LV load forecasting.
\end{abstract}

\vspace{0.35cm}

\end{@twocolumnfalse} 
] 



\section{Introduction}\label{sec:Introduction}
The increase of \ac{PV} systems, heat pumps and \acp{EV} implies challenges and new requirements for the planning and operation of distribution grids \cite{cakmakUsingOpenData2022}. \Acfp{DSO} have to cope with increased load and generation which results in higher peaks of active power. In order to prevent congestions for example due to thermal overload of \ac{DSO} equipment it is essential for the \ac{DSO} to know the loads induced by consumption and generation in the electric grid \cite{habenCoreConceptsMethods2023}. 

\begin{figure}
    \centering
    \includegraphics[width=0.25\textwidth]{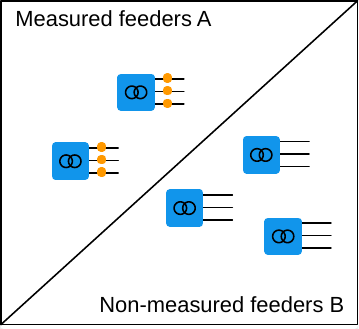}
    \caption{Classification of \ac{LV} feeders into measured ($A$) and non-measured feeders ($B$). The symbols represent substations with three \ac{LV} feeders. The present paper aims to estimate pseudo-measurements for non-measured feeders $B$ by learning the relation between measurements of feeders in $A$ and the feeder metadata like the number of houses or the installed power of heat pumps.}
    \label{fig:problem_statement}
\end{figure}

However, we observe that the rollout of measurement devices for \ac{DSO} equipment like \acf{LV} feeders is expensive and time-consuming. Consequently, the \ac{LV} feeders of \acp{DSO} are often divided into a set $A$ of measured feeders and a set $B$ of non-measured feeders as it is depicted in \Cref{fig:problem_statement}. 

For closing the measurement gap, it can be beneficial to estimate the active power of non-measured \ac{LV} feeders and substations. The estimated power values are subsequently named pseudo-measurements. In the present paper, we propose an approach based on feeder metadata for example the installed power of heat pumps or \ac{PV} systems along with the available measurements. A model is trained to learn how a measured power curve of a \ac{LV} feeder is related to the feeder metadata, the current timestamp and the weather conditions. During application, the model can predict the power of a non-measured feeder (a feeder from set $B$ in \Cref{fig:problem_statement}) with information about its metadata and the current timestamp and weather conditions. This approach does not require any measurements from the feeder for which the pseudo-measurements are generated. As long as the grid is not meshed, this pseudo-measurement approach is also applicable at substations by leveraging substation metadata. 

Pseudo-measurements can serve as a basis for multiple use cases in planning and operation processes. For planning, they indicate the remaining capacity in the feeder and facilitate the integration of new consumers and producers. If the feeder capacity is too low, an installation of measurement devices for more accurate monitoring or a grid expansion can be triggered. Moreover, the energy transition requires studies for the long-term planning of the grid in view of increasing load and generation. In Germany, \acp{DSO} are obligated to plan their grid with respect to climate neutrality in the year 2045 \cite{deutscherbundestagEnWG2024}. For this purpose, it is important for the \acp{DSO} to have knowledge about the current degree of capacity utilization. For operation, pseudo-measurements can improve the accuracy of a \ac{DSSE} \cite{dehghanpourSurveyStateEstimation2019}. Furthermore, the control of consumers and producers in case of grid congestions like thermal overload is becoming more important. In Germany, a new regulation allows the reduction of power consumption by devices in the \ac{LV} grid under specific circumstances \cite{bundesnetzagenturFestlegungZurDurchfuehrung2023}. Hereby, pseudo-measurements help to detect grid congestions and can be used to set up control schedules. 

Hence, in the present paper, we introduce a new approach to generate pseudo-measurements based on feeder metadata. We use a large real-world dataset from a \ac{DSO} in Southern Germany which includes metadata which is rare in the current literature \cite{habenReviewLowVoltage2021}. Furthermore, the measurement data is characterized by both load and feed-in. From $2{,}323$ feeders, $34{.}1$~\% exhibit at least ten times a net feed-in lower than $\num{-10}$~\si{\kW} mainly due to prosumers with \ac{PV} systems. Hence, we modify existing peak metrics to handle frequent zero-crossings as well as negative and positive peaks at one day.  

Our key contributions in the present paper are:
\begin{itemize}
    \item We present an approach to generate pseudo-measurements for \ac{LV} feeders based on feeder metadata.
    \item We evaluate the pseudo-measurements with respect to peak metrics inspired by the BigDEAL challenge \cite{shuklaBigDEALChallenge20222024} considering the magnitude, timing and shape of consumption and feed-in peaks.
\end{itemize}

The remainder of the paper is structured as follows. The related literature is presented in \Cref{sec:Related Work} and the problem statement is given in \Cref{sec:Problem statement}. The methodology in \Cref{sec:Methods} introduces the \ac{ML} approach of generating pseudo-measurements as well as the peak metrics. After introducing the dataset and the experiment in \Cref{sec:Experimental Setup}, we present the pseudo-measurements in \Cref{sec:Results}. Finally, we provide a discussion in \Cref{sec:Discussion}, the limitations in \Cref{sec:Limitations} and a conclusion in \Cref{sec:Conclusion}.

\section{Related Work}\label{sec:Related Work}
In this section, we first give an overview on different approaches to estimate the load and generation of a non-measured feeder in the \ac{LV} grid. Furthermore, we delimit the problem from adjacent research areas.

The lack of measurements in the grid is a common problem for \acp{DSO} and researchers focusing on distribution grids. \Cref{tab:related_work_transparency_approaches} shows different approaches to estimate the load. In particular, pseudo-measurements are essential for \ac{DSSE} if the grid is unobservable \cite{sanchezObservabilityLowVoltage2017}. But as emphasized in \Cref{sec:Introduction}, there are also other \ac{DSO} use cases for pseudo-measurements of \ac{LV} feeders besides \ac{DSSE} such as detecting a congestion at the feeder. 

The first approach to generate pseudo-measurements in \Cref{tab:related_work_transparency_approaches} is the \textit{smart meter exploitation} in the downstream grid. In case of an (almost) complete coverage by and availability of \acf{AMI}, the measurements at the grid connection customers can be aggregated to the feeder level (bottom-up) \cite{arrittComparingLoadEstimation2013, zhaoRobustMediumVoltageDistribution2020}. Furthermore, information from smart meters can be used to adjust load profiles \cite{krsmanPreprocessingPseudoMeasurements2016} or can be included directly in the \ac{DSSE} optimization problem \cite{rousseauxNewFormulationState2015}. However, the smart meter exploitation approach can be complex due to millions of distributed smart meters. More important, the data is often not available due to slow smart meter rollouts \cite{vitielloSmartMeteringRollOut2022}. This also applies for the grid area used in the present paper, which is why smart meter data is not used. Furthermore, legal and technical constraints can be a reason for restricted access of \acp{DSO} to smart meter data \cite{kroenerStateoftheArtIntegrationDecentralized2020}. 

Inversely, the second approach is the \textit{measurement disaggregation} from the upstream grid (top-down). A simple approach is to use ratios given by the \si{\kilovoltampere} rating of the transformers to distribute the measured load of \ac{MV} feeders \cite{arrittComparingLoadEstimation2013}. A disadvantage is the inaccuracy of the estimates, depending on the data used for building ratios. Disaggregation approaches are related to \ac{NILM} on the household level \cite{schirmerNonIntrusiveLoadMonitoring2023} which can also be extended to the grid level \cite{wangRegionalNonintrusiveLoad2020}. However, \ac{NILM} on the grid level aims to detect the contribution of equipment described in the feeder metadata, for example the proportion of power generated by \ac{PV} systems. This is not the aim in the present paper. 

The third approach \textit{explicit modeling} is used in \cite{rankovicModelingPhotovoltaicWind2012,morsyStateEstimationExogenous2020} to build pseudo-measurements for wind and solar generation. If domain knowledge about the generation of customers in the downstream grid is given, it is possible to model the estimated load by physical equations. Regarding the consumption of grid connection customers, a detailed bottom-up approach which combines the existing electrical devices with human activity patterns can be used \cite{proedrouComprehensiveReviewResidential2021}. The resulting generation and load profiles can be aggregated to the higher grid level (bottom-up). The main disadvantage of this approach is the complexity of modeling, especially with diverse customers such as different types of commerce and industry.

We refer to the fourth approach as \textit{synthetic load profiles} which can be aggregated to the higher grid level (bottom-up). In contrast to the third approach, the referenced methods need no detailed information about human activity or physical equations. Unlike the first approach, the load and generation profiles can be built with representative data from any downstream grid and we do not need \ac{AMI} in the downstream grid of the non-measured feeder. In Germany, the load profiles provided by the \ac{BDEW} are often used \cite{vdewReprasentativeVDEWLastprofile1999}. However, they are criticized for being imprecise \cite{grosParametrizationStochasticLoad2017}. More advanced examples comprise a \ac{VAE} called Faraday which can provide synthetic household profiles conditioned on consumption devices like \acp{EV} \cite{chaiFaradaySyntheticSmart2024}. Nevertheless, the aggregation of load profiles does not incorporate the simultaneity factors of customers and devices in the downstream \ac{LV} grid.

In contrast to the aforementioned approaches, the fifth approach consists in directly \textit{leveraging grid measurements} of feeders which are representative for the non-measured feeder (same grid level). The study presented in the present paper belongs to this fifth approach. We refer to the required data about the equipment in the \ac{LV} grid (see \Cref{tab:related_work_transparency_approaches}) as feeder metadata. In \cite{adinolfiPseudomeasurementsModelingUsing2014}, the feeder metadata comprises information about the residential and commercial customers and the public lighting. The authors in \cite{salazarDataDrivenFramework2020} frame the problem of generating load profiles for non-measured \ac{MV}/\ac{LV} transformers as transductive transfer learning. Despite solar irradiance, social demographic data and the installed power of \ac{PV} systems, they include load data from a similar transformer after clustering. 
Closest to the present paper, the authors in  \cite{ivasenkoNetzzustandsschatzungUndLeistungsvorhersage2022} propose a method based on substation metadata, weather data and calendar data to estimate load of non-measured substations which can also contribute to \ac{DSSE} \cite{keller-giessbachPotenzialKuenstlicherIntelligenz2020}. However, to the best of our knowledge, a detailed evaluation in literature is not yet available. 

In general, it can be observed that using feeder metadata is rare in the literature, also due to missing public datasets \cite{habenReviewLowVoltage2021}. If feeder metadata is used, it is often not detailed, for example including only the number of households without knowledge about large consumption devices like the number of \ac{EV} charging stations. However, such information can be crucial for the resulting load at the feeder level \cite{vasileAbschlussberichtNETZlaborEMobilityChaussee2022}. 

Other research areas related to pseudo-measurements cover the generation of synthetic time series and load forecasting. For synthetic time series, the difference is that they are not for specific feeders but should include specific characteristics, for example anomalies \cite{turowskiModelingGeneratingSynthetic2022, turowskiGeneratingSyntheticEnergy2024}. Load forecasting, including the emerging \acp{TSFM}, require measurements from the feeder itself \cite{habenCoreConceptsMethods2023, liangFoundationModelsTime2024}. However, measurements from the feeder are only available for a fraction of feeders in the present paper. Nevertheless, load forecasting is a very active research field and methods on time-series problems can often be adapted to regression problems like the one in the present paper \cite{habenReviewLowVoltage2021}. For example, the generative load forecasting model conditional Invertible Neural Network can also be used to generate pseudo-measurements based on feeder metadata \cite{heidrichUsingConditionalInvertible2024, treutleinErzeugungPseudomessdatenFur2023}. 

In literature about \ac{DSSE}, historical load data of a \ac{LV} feeder is often given, but recent or future load data is missing \cite{dehghanpourSurveyStateEstimation2019, carcangiuForecastingAidedMonitoringDistribution2020}. Generated recent or future load estimates of the \ac{LV} feeders (and other grid levels) are also called pseudo-measurements. However, analogous to load forecasting, the methods need historical data of the feeder itself. Hence, they are not suited for the problem statement in the present paper.

In addition to the methods, the tailored evaluation of load estimations plays a crucial role in distribution grids. Since \acp{DSO} need to plan and operate the grid with respect to the highest loads, evaluating with respect to the peak values is becoming more important. The BigDEAL challenge \cite{shuklaBigDEALChallenge20222024} in the field of load forecasting addresses this requirement as well as the authors in \cite{zuffereyImpactDataAvailability2021} for \ac{DSSE}. In the present paper, we adapt the peak metrics of \cite{shuklaBigDEALChallenge20222024} considering peak magnitude, timing and shape to cope with feeders exhibiting both consumption and feed-in. 

All in all, we observe that feeder metadata together with feeder measurements is rarely used in the literature to generate pseudo-measurements. Also, pseudo-measurements are mainly evaluated within a \ac{DSSE} but not isolated. The widely used load profiles are still essential due to missing measurements. However, they induce simplifications, for example by neglecting weather effects and could benefit from \ac{ML} techniques. Therefore, we propose an \ac{ML} method generating high resolution pseudo-measurements based on feeder metadata. The results are evaluated on a large real-world dataset with three models and nine metrics with a special focus on daily peak values. 

\begin{table*}[t]
    \centering
    \caption{Basic approaches in the literature which can be used to estimate the load and generation of non-measured \ac{LV} feeders and other grid levels.\\ Abbreviations: \acf{AMI}, \acf{LV}}
    \begin{tabular*}{\textwidth}{rlllrr}
     \toprule
     \# & Approach & Type & Important data requirement & Study examples & Reviews\\
     \midrule
     1 & \makecell[tl]{Smart meter\\ exploitation} & Bottom-up & \acs{AMI} in downstream grid  & \cite{arrittComparingLoadEstimation2013, rousseauxNewFormulationState2015, krsmanPreprocessingPseudoMeasurements2016, zhaoRobustMediumVoltageDistribution2020} & \cite{dehghanpourSurveyStateEstimation2019} \\ 
     2 & \makecell[tl]{Measurement\\ disaggregation} & Top-down & \acs{AMI} in upstream grid & \cite{arrittComparingLoadEstimation2013} & \\
     3 & \makecell[lt]{Explicit\\ modeling} & Bottom-up & \makecell[tl]{Deep domain knowledge\\ + data about equipment in \ac{LV} grid} & \cite{rankovicModelingPhotovoltaicWind2012, morsyStateEstimationExogenous2020} & \cite{proedrouComprehensiveReviewResidential2021}\\
     4 & \makecell[tl]{Synthetic load\\ profiles} & Bottom-up & \makecell[tl]{Representative measurements of grid connection customers\\ in any downstream grid + data about equipment in \ac{LV} grid} & \cite{vdewReprasentativeVDEWLastprofile1999, singhDistributionSystemState2010, chaiFaradaySyntheticSmart2024} & \cite{dehghanpourSurveyStateEstimation2019, proedrouComprehensiveReviewResidential2021} \\
     5 & \makecell[lt]{Leveraging grid\\ measurements} & \makecell[lt]{Same grid\\ level} & \makecell[lt]{Representative feeder measurements in any grid\\ + data about equipment in \ac{LV} grid} & \cite{arrittComparingLoadEstimation2013, adinolfiPseudomeasurementsModelingUsing2014, salazarDataDrivenFramework2020, ivasenkoNetzzustandsschatzungUndLeistungsvorhersage2022} &\\
     \bottomrule
    \end{tabular*}
\label{tab:related_work_transparency_approaches}
\end{table*}

\section{Problem statement}\label{sec:Problem statement}
\Cref{fig:problem_statement} illustrates a basic problem for \acp{DSO} in the \ac{LV} grid. Even though an increasing number of \ac{LV} feeders are equipped with measurement devices, there are still many feeders without any measured values. Therefore, our aim is to estimate pseudo-measurements for the non-measured feeders. It is important that the feeders in $A$ are representative for the whole grid ($A \cup B$) so that the estimation for the whole grid delivers reliable pseudo-measurements.

We divide the \ac{LV} feeders into a set $A$ and a set $B$ depending on the existence of measurements. Feeders in set $A$ can be used for training, whereas during inference, the unknown active power is estimated for feeders in $B$. The aim of the model training is to learn the parameters $\theta$ of the function $f$ with

\begin{equation} 
\hat{y}_{ij} = f_{\theta}(M_{ij}, W_{ij}, C_{j}),
\end{equation}

where $M_{ij} \in \mathbb{R}_{+}^k$ is the metadata of the feeder $i$ for timestamp $j$, $W_{ij} \in \mathbb{R}^l$ is the weather data related to the feeder $i$ for timestamp $j$ and $C_{j} \in \{0{,}1\}^m$ is calendar data for timestamp $j$. The variables $k$, $l$ and $m$ represent the dimension of the respective vector. The pseudo-measurement is denoted with $\hat{y}_{ij}$. 

The metadata $M_{ij}$ of a feeder $i$ change rarely compared to the weather data $W_{ij}$. Reasons for changes are mostly new registered equipment in the downstream grid of $i$. The measurement period can differ significantly between different feeders $i$ due to a continuous rollout of measurement devices. Furthermore, the measurements in the measured period can exhibit measuring gaps. We denote the set of timestamp indices where measurements for a feeder $i$ are existing with $N_i$.

A given function $f_{\theta}$ not only allows creating pseudo-measurements for the past and current timestamps. A forecast of the measurement can be received by passing predicted weather data $W_{ij}$ to $f_{\theta}$. Furthermore, different mid-term (months - few years) and long-term (many years) scenarios with varying metadata $M_{ij}$ or weather data $W_{ij}$ can be conducted, for example to simulate new installations and ramp-ups of electrical equipment.

\section{Feeder metadata-based pseudo-measurements}\label{sec:Methods}
This section describes the methodology to generate pseudo-measurements for feeders without measurements based on feeder metadata. It includes data requirements and the model pipeline which is visualized in \Cref{fig:model_architecture}. Furthermore, the peak metrics which are adapted from the BigDEAL challenge \cite{shuklaBigDEALChallenge20222024} are introduced. 

\subsection{Data requirements}
The required data to train and operate a pseudo-measurement model based on metadata comprises different data sources. In order to provide a target variable for the model training, a sufficiently large amount of measured \ac{LV} feeders in the grid is needed.   

The essential feature data for our proposed model is metadata collected by the \ac{DSO}. We divide the feeder metadata into three groups. 

The number of housing units or the number of commerce and industry units belong to the first group of grid connection points. The second group comprises installed power of equipment which describes producers like \ac{PV} systems or \ac{CHP} and consumers like heat pumps or household units. It is necessary that the equipment is allocated to the respective \ac{LV} feeder and is registered with the installed electrical power. The third group of feeder metadata provides information about the energy consumption. Note that energy consumption data does not mean that the grid node has measurements. Rather, it is highly aggregated consumption data, for example annual consumption of grid connection points in the \ac{LV} grid. This can improve the model because the grid connection points or installed power of equipment can differ based on different usage behavior of houses and industries at the \ac{LV} feeder. The consumption data is used as a feature, therefore it needs to be known for both measured (training) and non-measured (inference) feeders. 

In addition to feeder metadata, appropriate weather data should be provided since generation and consumption of installed equipment depend highly on the weather. Examples of relevant variables include temperature in case of heating equipment or global radiation in case of \ac{PV} systems. 

Moreover, calendar information about holidays or other events can improve the model accuracy, because the behavior of people differs between workdays, weekends and holidays. Finally, a timestamp encoding is needed depending on the required temporal resolution of the pseudo-measurements. 

\subsection{Model pipeline}
\begin{figure*}
    \centering
    \includegraphics[width=1\textwidth]{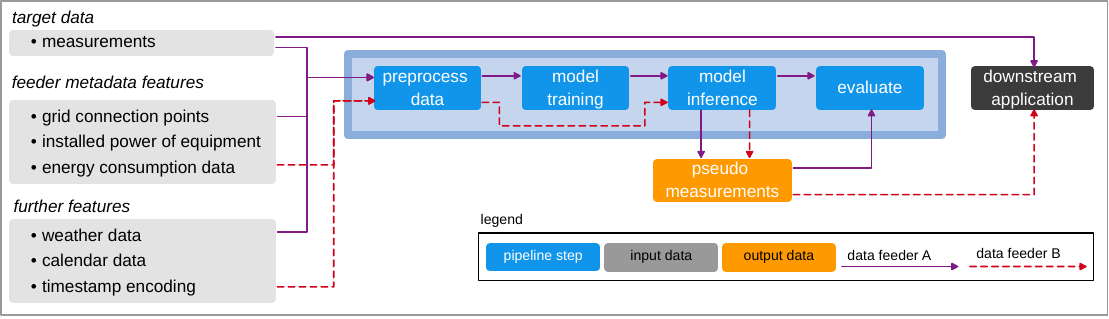}
    \caption{Model pipeline with the pipeline steps, input data and output data. The data flows are differentiated with respect to measured feeders $A$ and non-measured feeders $B$ as described in \Cref{sec:Problem statement}.}
    \label{fig:model_architecture}
\end{figure*}
\Cref{fig:model_architecture} illustrates the model pipeline with the different pipeline steps and the most important input and output data. First, the feature and target data are preprocessed. Due to the diverse data sources, the preprocessing includes many subroutines such as aligning the different time resolutions between groups of data or aggregating several metadata features to one feature. The second step is the model training using the data of all feeders in $A$, where different \ac{ML} regression models for tabular data can be applied. In contrast to \ac{LV} load forecasting, the model does not use autoregressive features of the feeder because they are not available for feeders in $B$. Afterward, the pseudo-measurements are generated for all feeders in $A$ and $B$. The pseudo-measurements of feeders in $A$ are evaluated with respect to the existing ground truth data.\footnote{We neglect that feeders in $A$ are divided into training, validation and test data so that the model can be evaluated on unseen data. Refer to \Cref{sec:Experimental Setup} for more details.} The pseudo-measurements of feeders in $B$ are used together with the real measurements of feeders in $A$ in the downstream application of the \ac{DSO}.

There is no need to pass an ID of the \ac{LV} feeder to the model, because the feeder is defined by the characteristics of its metadata. Furthermore, the model does not forecast a time-series, because the feeders are in general not measured and there is no data available. However, if the features represent future meta- and weather data, the architecture can also be used to forecast.  

\subsection{Metrics for evaluation}
In order to evaluate the pseudo-measurement model for individual feeders, we use the following nine metrics which are grouped into three all-observation metrics and six peak metrics.

Let $y_{ij}$ be the measured value of feeder $i$ at timestamp $j$ and $\hat{y}_{ij}$ the respective estimated pseudo-measurement. We denote the \ac{MAE} with
\begin{equation}
\text{MAE}_i = \frac{1}{|N_i|} \sum_{j \in N_i} \mid y_{ij} - \hat{y}_{ij} \mid
\end{equation}
and the \ac{RMSE} with
\begin{equation}
\text{RMSE}_i = \sqrt{\frac{1}{|N_i|} \sum_{j \in N_i} (y_{ij} - \hat{y}_{ij})^2}
\end{equation}
as \textit{all-observation} metrics because the model is evaluated against all existing measurements of a feeder. Hereby, $N_i$ is the set of timestamp indices of a feeder $i$ where measurements are available and $|N_i|$ the cardinality. Furthermore, we denote 
\begin{equation}
    \text{MAE}_{i, norm} = \frac{\text{MAE}_i}{\max\{y_{ij} \mid j \in N_i\} - \min\{y_{ij} \mid j \in N_i\}}
\end{equation}
as the \ac{MAE} normalized by the min-max range of the feeder $i$. 

Additional to the all-observation metrics, we introduce the metrics focusing on peaks. The metrics are inspired by the BigDEAL forecasting challenge and incorporate special requirements of \acp{DSO} \cite{shuklaBigDEALChallenge20222024} facing both consumption and feed-in at the \ac{LV} feeders. The basic metrics are \ac{PMag} with unit \si{\kW}, \ac{PTime} with unit hour (h) and \ac{PShape} without unit. 

Compared to \cite{shuklaBigDEALChallenge20222024}, we make five adaptations in total. This is necessary, because unlike in \cite{shuklaBigDEALChallenge20222024}, the data is characterized by both consumption and feed-in and many \ac{LV} feeders exhibit zero-crossing measurements. 

First, the metrics are evaluated for both positive and negative peaks to consider the consumption and feed-in estimation. More precisely, the consumption and feed-in is the resulting net consumption and net feed-in at the feeder after the prosumption of all grid connection customers belonging to the feeder. Second, two thresholds are introduced so that only feeders with at least ten daily peaks exceeding $\pm 10$~\si{\kW} are evaluated. Thereby, days and feeders with low consumption and feed-in are excluded. Third, the base metric of \ac{PMag} is changed from \ac{MAPE} to \ac{MAE} because the data exhibits many zero-crossings which make \ac{MAPE} unsuitable. Fourth, the base metric of \ac{PTime} is changed from a piecewise linear function to \ac{MAE} to facilitate the interpretation and preserve the unit hour. Fifth, we use a min-max normalization for \ac{PShape} instead of a peak normalization, again to account for the frequent zero-crossings in the measurements. 

We introduce the sets of pairs $P_i^{C}$ and $P_i^{F}$ to describe the timestamp indices of the peaks for the daily consumption (C) and the daily feed-in (F). For a feeder $i$, $P_i^{C} = \emptyset$ if there are less than $10$ days where the measurement exceeds $+10$~\si{\kW}. Otherwise, a tuple $(j_1, j_2) \in P_i^{C}$ for a day $d$ if the measurements are greater or equal $+10$~\si{\kW} at least once on $d$. The first tuple value $j_1$ is the timestamp index of the maximum in the measurement (ground truth) on day $d$. The second tuple value $j_2$ is the timestamp index of the pseudo-measurement maximum on day $d$. Analogously, we define $P_i^{F}$ for feed-in peaks lower or equal $-10$~\si{\kW}. 

Hence, we denote
\begin{equation} 
    \text{PMag}_i^{T} = \frac{1}{|P_i^{T}|} \sum_{(j_1{,}j_2) \in P_i^{T}} |y_{ij_1} - \hat{y}_{ij_2}|
\end{equation}
and
\begin{equation} 
    \text{PTime}_i^{T} = \frac{1}{|P_i^{T}|} \sum_{(j_1{,}j_2) \in P_i^{T}} |j_1 - j_2|
\end{equation}
where $T \in \{C{,}F\}$ is the peak type, $|P_i^{T}|$ is the cardinality and $|j_1 - j_2|$ the absolute difference between two timestamps in hours.  

Finally, we define the \ac{PShape} with
\begin{equation} 
    \text{PShape}_i^{T} = \frac{1}{|P_i^{T}|} \sum_{(j_1{,}j_2) \in P_i^{T}} \sum_{x \in Q} |s_{ix} - \hat{s}_{ix}|
\end{equation}
where $Q = \{x \mid x \in (j_1 - 2, ..., j_1 + 2)\}$ is the set of timestamps two hours around the peak in the measurement $j_1$ (ground truth) with 

\begin{equation}
    s_{ix} = \frac{y_{ix} - \min\{y_{ij} \mid j \in Q\}}{\max\{y_{ij} \mid j \in Q\} - \min\{y_{ij} \mid j \in Q\}}
\end{equation}
and 
\begin{equation}
    \hat{s}_{ix} = \frac{\hat{y}_{ix} - \min\{\hat{y}_{ij} \mid j \in Q\}}{\max\{\hat{y}_{ij} \mid j \in Q\} - \min\{\hat{y}_{ij} \mid j \in Q\}}.
\end{equation}
representing the (pseudo) measurement min-max normalized around the peak.

\section{Experimental Setup}\label{sec:Experimental Setup}
First, this section describes the data set including the data cleaning and preparation steps. Second, we describe the structure of the conducted experiments. 

\subsection{Data set and feature engineering}\label{subsec:experimental_setup-data_set}
\begin{table}[]
    \caption{Different characteristics of the dataset used in the experiments after applying the data cleaning and the feature engineering specified in \Cref{subsec:experimental_setup-data_set}.}
    \centering
    \begin{tabular}{lr}
        Characteristic & Value\\
        \toprule
        Number of feeders & 2,323 \\
        Number of measurements & 52,039,067\\
        Number of metadata categories & 21\\
        \midrule
        Avg. measurements per feeder & 22,402\\
        Lowest measurement & -146.69~\si{\kW} \\ 
        Highest measurement & 178.59~\si{\kW}\\
        \midrule
        First measurement & Feb 23, 2022\\
        Last measurement & March 28, 2024\\
        Avg. measurement period per feeder & 233 days\\
        \bottomrule
    \end{tabular}
    \label{tab:dataset_table}
\end{table}
We use active power measurements of \ac{LV} feeders together with metadata from the \ac{DSO} \textit{Netze BW} in the southwest of Germany. This data is not open due to privacy regulations. The weather data originates from the \ac{DWD} \cite{reinertDWDDatabaseReference2024}. 

\Cref{tab:dataset_table} summarizes basic characteristics of the dataset with $2{,}323$ \ac{LV} feeders. The approximately $52$ million measurements in $15$ minute resolution represent an aggregation from minute resolution data by mean. The measurement period ranges from February 23, 2022 to March 28, 2024 with different starting points for feeders due to the continuous rollout. Therefore, a feeder has on average $22{,}402$ measurements which is around $233$~days. Meshed feeders are not included in the dataset. 

We use filters in order to detect feeders with implausible data which results in the $2{,}323$ feeders after applying. Regarding the measurements, a feeder needs in total at least one day of measurements and the measurement must exceed $\pm 5$~\si{\kW} at least one time. Regarding the feeder metadata, feeders are removed from the dataset where the installed power of a single consumption or generation category exceeds the physical limit of the feeder. Furthermore, feeders without metadata are removed as well as feeders where the feed-in exceeds the installed power of producers by more than $5$~\si{\kW}. If a feed-in is observed between midnight and two a.m., but the only producer are \ac{PV} systems, the feeder is also removed. 

\begin{figure}
    \centering
    \includegraphics[width=1\columnwidth]{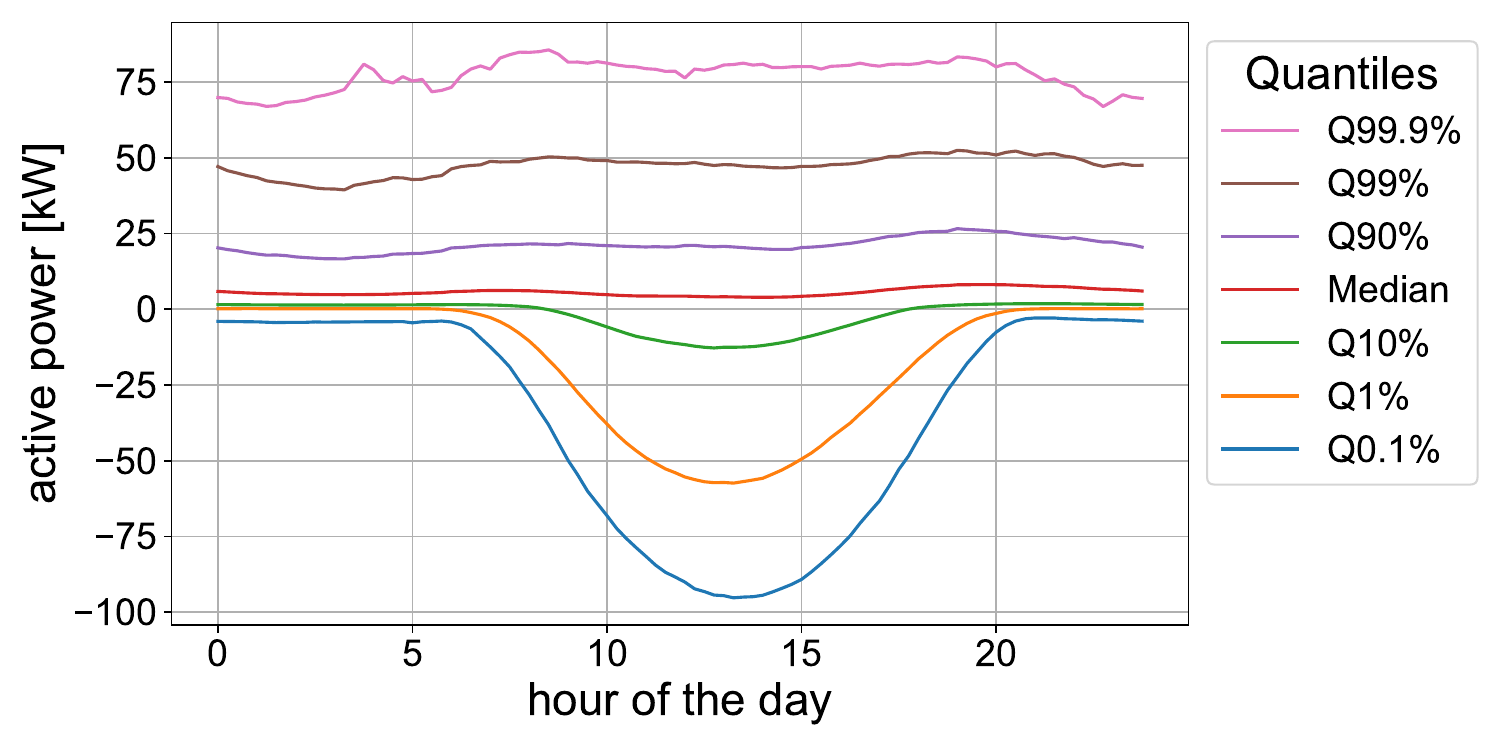}
    \caption{Daily profile of the measurements from all feeders for different quantiles in $15$ minute resolution after applying the data cleaning and feature engineering specified in \Cref{subsec:experimental_setup-data_set}.}
    \label{fig:quantile_measurement_plot}
\end{figure}
\Cref{fig:quantile_measurement_plot} shows different quantiles of the measurements for all feeders in a daily load profile. It is visible that both consumption and feed-in during the day are present. The median ranges between $0$~\si{\kW} and $10$~\si{\kW}, whereas the $90$~\%, $99$~\% and $99.9$~\% quantiles are close to $25$~\si{\kW}, $50$~\si{\kW} and $75$~\si{\kW} respectively. The feed-in shows a \ac{PV} curve, where the $0.1$~\% quantile is close to $-100$~\si{\kW}. \Cref{fig:violinplot_measurements} shows the monthly distribution. All months exhibit long tails for both consumption and feed-in. The distribution in the winter is  broader compared to the summer. The highest consumption occurs in the winter with a maximum of $178.59$~\si{\kW}. The highest feed-in is in spring and summer with a minimum of $-146.69$~\si{\kW}. 

We do not include an outlier detection and removal of measurements, since the measurements seem reasonable. Moreover, correctly predicting high values is essential for \acp{DSO} and outlier removal involves the risk of removing extreme but correct data. 

Before the data is fed into the model, we aggregate similar feeder metadata like different types of \textit{heat pumps}. In total, this results in $21$ categories for feeder metadata summarized in \Cref{tab:metadata_features_table}. We include a feature for the number of \textit{housing units} which belongs to the group of grid connection points. Regarding the installed power of equipment, there are nine features for consumers, one for \textit{batteries} and two for producers. As energy consumption data, we include eight features for different types of commerce and industry named $g0$ - $g6$ and $l0$. They correspond to the customer groups defined in \cite{vdewReprasentativeVDEWLastprofile1999}. The features are derived from monthly or yearly billing data and describe the average energy consumption per day in \si{\kWh}. The yearly billing data of households is not given in the metadata. 

\begin{figure*}
    \centering
    \includegraphics[width=1\textwidth]{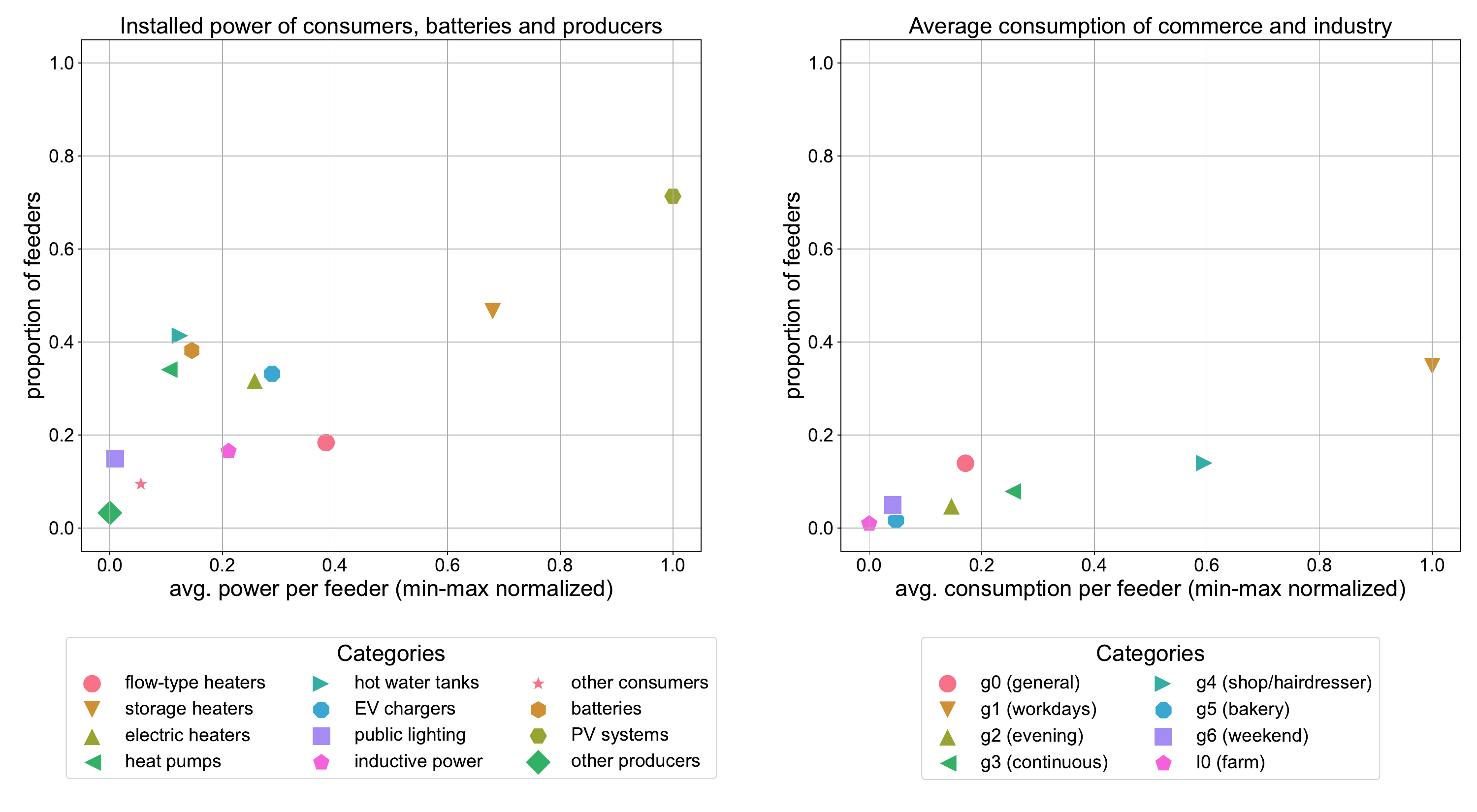}
    \caption{Distribution of metadata at \ac{LV} feeders divided into consumers, batteries and producers (left) as well as the average consumption data of commerce and industry (right). The x-axis is scaled based on the categories \textit{solar} and \textit{g1} respectively. The y-axis describes the fraction of feeders where the category is greater than zero.}
    \label{fig:meta_data_distribution}
\end{figure*}
All of the feeder metadata categories except the number of \textit{housing units} are visualized in \Cref{fig:meta_data_distribution}. The left part shows the \textit{installed power of equipment}. \textit{\ac{PV} systems} are installed at $70$~\% of all feeders and show the highest average installed power per feeder. Some categories such as \textit{batteries} are present at many feeders but with relatively low installed power, other categories like \textit{flow-type heaters} have reverse characteristics. The right part shows the average \textit{energy consumption data} of commerce and industry. The category \textit{g1} for industry which produces mainly on workdays during 8 a.m. to 6 p.m. is dominating. The \textit{housing units} are the only category from the feeder metadata group \textit{grid connection points} as defined in \Cref{fig:model_architecture}. 

Regarding the weather we include four features, namely the global radiation, air temperature, precipitation and the snow height.  

In order to model the timestamp information $j$, we encode the day of the year, the day of the week and the minute of the day with a cyclical encoding using a sine and a cosine function. This results in six features describing the timestamp. We further include two binary features describing if the day is a holiday or workday.

In total, we use $33$ features which cover various effects on the target variable representing the active power measurement in \si{\kW} at the \ac{LV} feeder. 

\subsection{Conducted experiments}
\begin{figure}
    \centering
    \includegraphics[width=1\columnwidth]{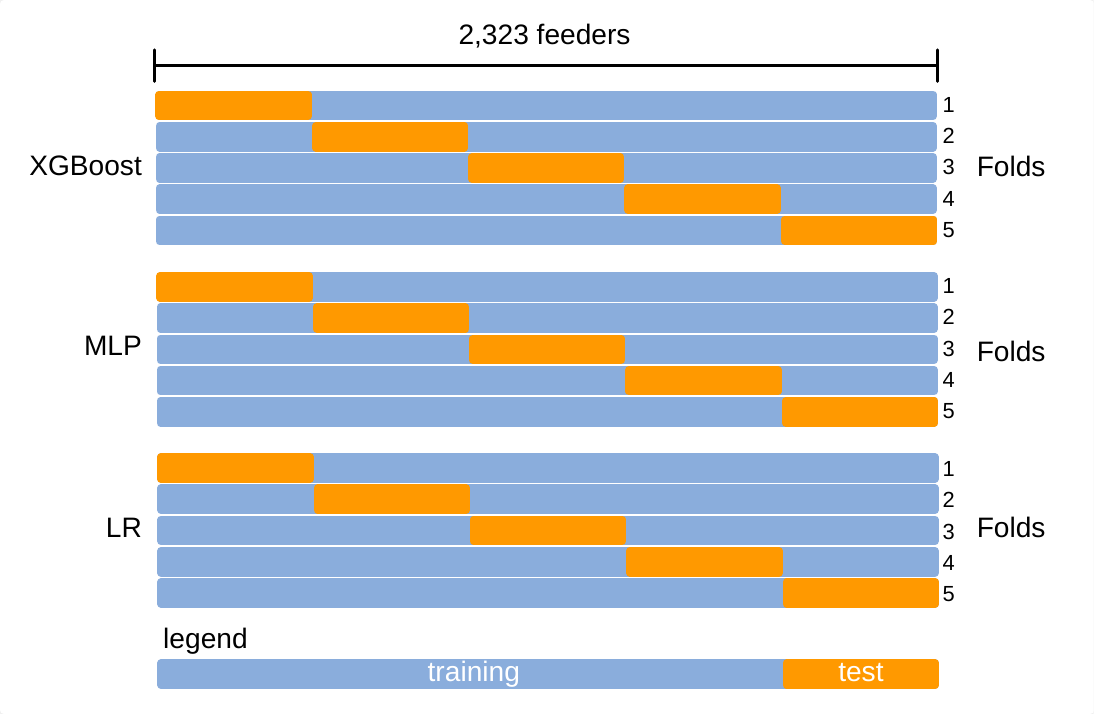}
    \caption{Visualization of the used cross-validation.}
    \label{fig:cross_validation_vissualization}
\end{figure}
The generation of pseudo-measurements is evaluated with three models based on an experiment with a $5$-fold cross-validation. This results in $15$ sub-experiments with $15$ different models as shown in \Cref{fig:cross_validation_vissualization}. The cross-validation makes it possible to evaluate the method for each model on all feeders, because every feeder is exactly once in the test data. 

In each of the $15$ sub-experiments, we use a train-test split of the feeders with a ratio of $80$~\% to $20$~\%. Applied to \Cref{fig:problem_statement}, the $80$~\% represent the feeder set $A$ and the $20$~\% represent $B$. Within one sub-experiment, the data of one feeder can be either in the training or in the test data, but not both. We do not split after time as it is common in load forecasting, because the primary problem statement is to estimate values of non-measured feeders and not future values of the same feeder. 

All metrics and visualizations reported in \Cref{sec:Results} are solely based on feeders in test data. Regarding \Cref{fig:problem_statement}, all of the feeders used in the experiments belong to the set $A$. With the retention of feeders in $A$ as test data, we can determine the expected performance of the pseudo-measurements for feeders in $B$.

The used models are \ac{LR} \cite{scikit-learn}, \ac{MLP} \cite{scikit-learn} and \ac{XGBoost} \cite{chenXGBoostScalableTree2016b}. We choose these models to include a linear model, a neural network and a tree-based model. Tree-based models like \ac{XGBoost} are state-of-the art for tabular regression problems like in the present paper \cite{grinsztajnWhyTreebasedModels2022}.

Regarding the \ac{XGBoost} and the \ac{MLP} model $12.5$~\% of the training data is used as validation data for early stopping of the training routine to prevent overfitting. The hyperparameters are chosen based on pre-experiments on similar data and can be seen in \Cref{tab:hyperparameters}. When using the \ac{MLP}, we apply a min-max normalization to the features and targets. The \ac{LR} is used with elastic net regularization. 

All experiments are conducted with Azure Machine Learning on a compute cluster with $16$ cores, $112$~\si{\giga\byte} RAM and $224$~\si{\giga\byte} disk. 

\section{Results}\label{sec:Results}

First, we present the all-observation metrics and the peak metrics evaluated for all cross-validation runs and all models. Afterwards, excerpts of pseudo-measurements for selected feeders are shown for the \ac{XGBoost} model of fold~$1$.  

\subsection{All-observation metrics}
\begin{table}[]
    \centering
    \caption{Metrics evaluated for all timestamps and all feeders (all-observation metrics). The metric is evaluated on the feeders in the test data of a cross-validation fold and the results of all folds are combined into this table. Best values of the three models are bold.}
    \begin{tabular}{llrrr}
        \toprule
        Metric & \makecell[l]{Statistical\\ measure} & XGBoost & MLP & LR \\
        \midrule
        \acs{MAE} [\si{\kW}] & mean & \textbf{4.73} & 4.85 & 6.07 \\
         & std & \textbf{3.35} & 3.78 & 4.01 \\
         & min & \textbf{0.70} & 0.76 & 1.24 \\
         & 25~\% & \textbf{2.47} & 2.51 & 3.71 \\
         & median & \textbf{3.83} & 3.84 & 4.84 \\
         & 75~\% & \textbf{5.83} & 5.90 & 7.16 \\
         & max & \textbf{33.98} & 64.51 & 90.55 \\
         \midrule
        \acs{MAE}$_{norm}$ & mean & \textbf{0.13} & \textbf{0.13} & 0.17 \\
         & std & 0.24 & 0.22 & \textbf{0.21} \\
         & min & \textbf{0.01} & 0.02 & 0.03 \\
         & 25~\% & \textbf{0.06} & 0.07 & 0.08 \\
         & median & \textbf{0.09} & \textbf{0.09} & 0.12 \\
         & 75~\% & \textbf{0.13} & 0.14 & 0.19 \\
         & max & 7.24 & 6.34 & \textbf{5.38} \\
         \midrule
        \acs{RMSE} [\si{\kW}] & mean & \textbf{6.19} & 6.33 & 7.85 \\
         & std & \textbf{4.35} & 4.77 & 5.19 \\
         & min & \textbf{0.88} & 0.96 & 1.79 \\
         & 25~\% & \textbf{3.17} & 3.25 & 4.62 \\
         & median & 5.09 & \textbf{5.05} & 6.12 \\
         & 75~\% & \textbf{7.86} & 7.89 & 9.51 \\
         & max & \textbf{45.41} & 70.80 & 91.02 \\
        \bottomrule
    \end{tabular}
    \label{tab:describe_all_observation_metrics}
\end{table}
\Cref{tab:describe_all_observation_metrics} includes different statistical measures for the metrics \ac{MAE}, \ac{MAE}$_{norm}$ and \ac{RMSE} evaluated for the models \ac{XGBoost}, \ac{MLP} and \ac{LR}. 

\textit{\ac{XGBoost} and \ac{MLP} show superior performance} compared to \ac{LR} for all three metrics. The average \ac{MAE} over all feeders for \ac{XGBoost} is $4.73$~\si{\kW} which is $1.34$~\si{\kW} better compared to $6.07$~\si{\kW} for \ac{LR}. The performance of \ac{XGBoost} is slightly better for the mean of the all-observation metrics compared to \ac{MLP}. 

\textit{All metric distributions for all models are heavily right skewed} which is stated by $median < mean$.  

\textit{The model performance differs between different train-test-splits} which is shown in \Cref{tab:fold_standard_deviations}. For example, the standard deviation of the median \ac{RMSE} regarding the five folds is between $0.22$~\si{\kW} and $0.49$~\si{\kW} for all models combined.

\subsection{Peak metrics}
\begin{table}[]
    \centering
    \caption{Peak metrics evaluated for all feeders and daily consumption and feed-in peaks. The metric is evaluated on the feeders in the test data of a cross-validation fold and the results of all folds are combined into this table. Best values of the three models are bold.}
    \begin{tabular}{llrrr}
        \toprule
        Metric & \makecell[l]{Statistical\\ measure} & XGBoost & MLP & LR \\
        \midrule
        $\text{PMag}^{C}$ [\si{\kW}] & count & 1921 & 1921 & 1921 \\
        (consumption) & mean & 11.66 & \textbf{11.51} & 14.06 \\
         & std & \textbf{9.09} & 9.46 & 10.76 \\
         & min & \textbf{0.76} & 0.98 & 0.80 \\
         & 25~\% & 5.85 & \textbf{5.59} & 6.57 \\
         & median & 9.02 & \textbf{8.55} & 11.24 \\
         & 75~\% & 14.48 & \textbf{14.35} & 18.39 \\
         & max & \textbf{88.02} & 117.71 & 117.08 \\
        \midrule
        $\text{PMag}^{F}$ [\si{\kW}] & count & 793 & 793 & 793 \\
        (feed-in) & mean & 13.13 & \textbf{12.52} & 21.36 \\
         & std & \textbf{7.50} & 7.68 & 13.24 \\
         & min & \textbf{1.06} & 1.81 & 1.59 \\
         & 25~\% & 7.98 & \textbf{7.32} & 11.83 \\
         & median & 11.44 & \textbf{10.78} & 18.54 \\
         & 75~\% & 16.60 & \textbf{15.14} & 28.10 \\
         & max & \textbf{55.56} & 67.21 & 73.71 \\
         \midrule
        $\text{PTime}^{C}$ [\si{\hour}] & count & 1921 & 1921 & 1921 \\
        (consumption) & mean & \textbf{5.07} & 5.14 & 8.02 \\
         & std & 2.82 & 2.80 & \textbf{1.39} \\
         & min & \textbf{0.54} & 0.56 & 2.10 \\
         & 25~\% & \textbf{2.97} & 3.03 & 7.16 \\
         & median & \textbf{4.40} & 4.45 & 7.94 \\
         & 75~\% & \textbf{6.54} & 6.67 & 8.90 \\
         & max & 16.33 & 16.57 & \textbf{13.35} \\
        \midrule
        $\text{PTime}^{F}$ [\si{\hour}] & count & 793 & 793 & 793 \\
        (feed-in) & mean & 1.17 & 1.27 & \textbf{1.13} \\
         & std & 0.69 & 0.81 & \textbf{0.43} \\
         & min & \textbf{0.50} & 0.50 & 0.54 \\
         & 25~\% & \textbf{0.95} & 0.97 & 0.97 \\
         & median & \textbf{1.05} & 1.09 & 1.08 \\
         & 75~\% & \textbf{1.19} & 1.29 & 1.21 \\
         & max & 12.04 & 11.24 & \textbf{9.15} \\
        \midrule
        $\text{PShape}^{C}$ & count & 1921 & 1921 & 1921 \\
        (consumption) & mean & 0.35 & \textbf{0.34} & 0.38 \\
         & std & 0.05 & 0.05 & \textbf{0.04} \\
         & min & 0.19 & \textbf{0.18} & 0.23 \\
         & 25~\% & 0.32 & \textbf{0.31} & 0.35 \\
         & median & 0.35 & \textbf{0.34} & 0.38 \\
         & 75~\% & 0.38 & \textbf{0.37} & 0.41 \\
         & max & 0.60 & \textbf{0.54} & 0.55 \\
        \midrule
        $\text{PShape}^{F}$ & count & 793 & 793 & 793 \\
        (feed-in) & mean & \textbf{0.29} & \textbf{0.29} & \textbf{0.29} \\
         & std & \textbf{0.02} & \textbf{0.02} & \textbf{0.02} \\
         & min & \textbf{0.23} & \textbf{0.23} & \textbf{0.23} \\
         & 25~\% & \textbf{0.28} & \textbf{0.28} & \textbf{0.28} \\
         & median & \textbf{0.29} & \textbf{0.29} & \textbf{0.29} \\
         & 75~\% & \textbf{0.30} & \textbf{0.30} & \textbf{0.30} \\
         & max & 0.40 & \textbf{0.37} & 0.39 \\
        \bottomrule
    \end{tabular}
    \label{tab:describe_peak_metrics}
\end{table}
In contrast to the all-observation metrics, the peak metrics are only evaluated for feeders with daily consumption and feed-in peaks exceeding $\pm 10$~\si{\kW} at least $10$ times. \Cref{tab:describe_peak_metrics} highlights the \ac{PMag}, \ac{PTime} and \ac{PShape} metric for the consumption ($1921$ feeders evaluated) and feed-in ($793$ feeders evaluated).

\textit{Analogous to the all-observation metrics, the \ac{PMag} for consumption and feed-in is heavily right skewed} throughout all models except for the feed-in \ac{PTime}. For example, the median of the $\text{PMag}^{C}$ for the \ac{MLP} is with $8.55$~\si{\kW} lower compared to the mean of $11.51$~\si{\kW}. 

\textit{The \ac{PMag} for both consumption and feed-in is significantly higher compared to the \ac{MAE}} for all models. While the median \ac{MAE} of the \ac{XGBoost} model is $3.83$~\si{\kW}, the median of the $\text{PMag}^{C}$ is $9.02$~\si{\kW} and the median of $\text{PMag}^{F}$ is $11.44$~\si{\kW}. This implies that the peak estimation of the models is worse than the estimation of the complete time-series. 

\textit{When comparing consumption and feed-in, the \ac{PMag} for consumption is lower compared to feed-in} with respect to the mean and the quantiles given in \Cref{tab:describe_peak_metrics} for all models. For example, the mean of \ac{PMag}$^{C}$ of \ac{XGBoost} is with $11.66$~\si{\kW} lower than the mean of \ac{PMag}$^{F}$ with $13.13$~\si{\kW}. 

\textit{\ac{PTime} is significantly better for feed-in compared to consumption}. While the mean of the \ac{PTime}$^{F}$ is $1.17$~\si{\hour} for \ac{XGBoost}, the mean of \ac{PTime}$^{C}$ is $5.07$~\si{\hour}.

\textit{\ac{PShape} is better for feed-in compared to consumption}. While the mean of the \ac{PShape}$^{F}$ is $0.29$ for \ac{XGBoost}, the mean of the \ac{PShape}$^{C}$ is $0.35$. 

\textit{The \ac{PShape}$^{F}$ shows smaller fluctuations compared to \ac{PShape}$^{C}$}. We observe for all models combined that the $25$~\% and $75$~\% quantile is between $0.28$ and $0.30$ for \ac{PShape}$^{F}$ and between $0.31$ and $0.41$ for \ac{PShape}$^{C}$.

\textit{The \ac{PMag} and \ac{PTime} reveal some feeders with very high deviations of the peaks in magnitude and time}. For example, the highest errors for \ac{PTime}$^{C}$ can be observed for feeders with storage heaters, exhibiting a peak in the night after midnight while the pseudo-measurement estimates the peak in the evening.


\subsection{Selected feeders}
\begin{figure*}[t]
    \centering
    \includegraphics[width=1\textwidth]{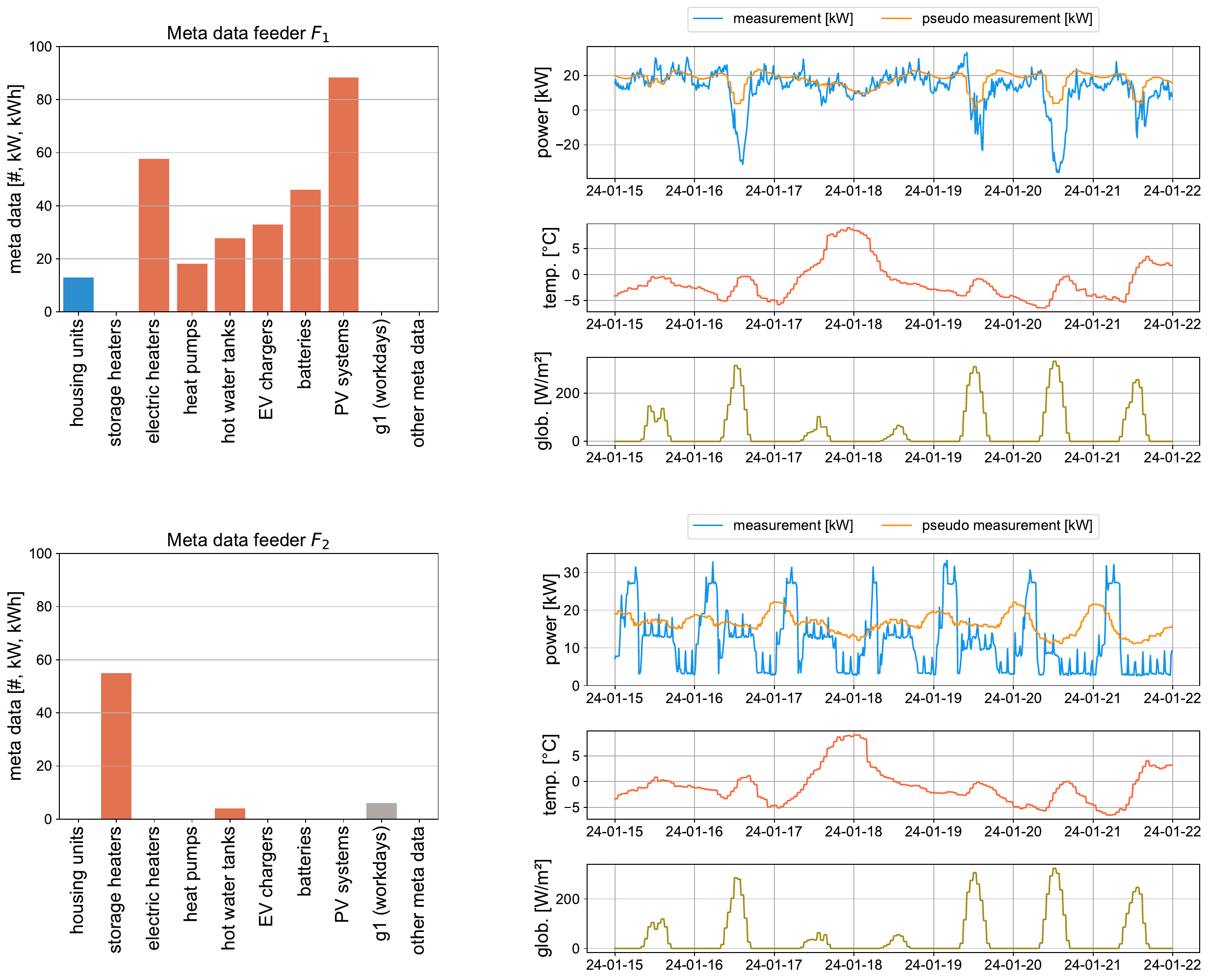}
    \caption{Two \ac{LV} feeders with the feeder metadata on the left side and the measurement and pseudo-measurement on the right side together with the temperature and global radiation of one week in January $2024$ (Monday - Sunday). The feeders represent the $10$~\% ($F_1$) and $90$~\% ($F_2$) quantile of the metric \ac{MAE}$_{norm}$ in the \ac{XGBoost} fold~$1$ in a pre-experiment. The metric values for the feeders are given in \Cref{tab:feeder_metric_details}.}
    \label{fig:one_feeder_plot}
\end{figure*}
\Cref{fig:one_feeder_plot} depicts two feeders $F_1$ and $F_2$ with the respective feeder metadata and an excerpt of one week for the measurements and generated pseudo-measurements as well as temperature and global radiation. The chosen week is a week in the winter of January 2024 with a temperature anomaly at Wednesday, January 17 to Thursday, January 18. The global radiation is between $0$ and $333$~\si{\frac{\watt}{\meter^2}}. $F_1$ represents the $10$~\% quantile and $F_2$ the $90$~\% quantile of the \ac{MAE}$_{norm}$ metric in a pre-experiment. More details are given in the appendix: \Cref{fig:quantile_plots} shows weekly quantile profiles for the feeders $F_1$ and $F_2$ and in addition for three other feeders $F_3$ - $F_5$. \Cref{tab:feeder_metric_details} includes the individual metrics for $F_1$ - $F_5$ with respect to the complete time-series of the feeders (not only the excerpt).

Feeder $F_1$ is characterized by $13$ housing units and diverse metadata including heating systems, \ac{EV} chargers, batteries and \ac{PV} systems. In the measurements, we observe a load between $-35.94$~\si{\kW} and $33.18$~\si{\kW} and feed-in for at least four days with power smaller than $0$~\si{\kW}. The lowest load during the night can be observed during the highest temperature of $9$~\si{\degreeCelsius} on January 17 and 18. Likewise, we observe that the pseudo-measurements are low in the evening of January 17 and the morning of January 18 not exceeding $20$~\si{\kW}. On January 19, we detect the highest load whereas the feed-in peak is less compared to January 16 and January 20 with similar global radiation around $300$~\si{\frac{\watt}{\meter^2}}. 

The \ac{MAE}$_{norm}$ of $0.04$ indicates a good fit of the base load and the \ac{PMag}$^{C}$ of $7.14$~\si{\kW} is low compared to the fold average of $11.78$~\si{\kW}. While the \ac{PTime}$^{C}$ of $7.38$~\si{\hour} is high compared to the average, the \ac{PShape}$^{C}$ of $0.32$ is low in comparison to the average performance. We recognize for January 19 and January 20 that the consumption peak in the measurement is in the morning and the pseudo-measurement consumption peak is in the evening.

The \ac{PMag}$^{F}$ of $18.46$~\si{\kW} is higher than the average \ac{PMag}$^{F}$ of \ac{XGBoost} fold~$1$ with $12.58$~\si{\kW}. We observe that the highest error for the feed-in peak estimation is on January 20 with $41.71$~\si{\kW} deviation at $3$~pm. The \ac{PTime}$^{F}$ and \ac{PShape}$^{F}$ of $1.10$~\si{\hour} and $0.29$ are close to the respective fold averages. 

Feeder $F_2$ is characterized by industry and commerce with storage heaters and a low installed power of hot water tanks. In the measurements, we observe only positive load values with high peaks during the night, an increased load during the working hours from Monday, January 15 to Friday, January 19 and regular load spikes every few hours. The load at the weekend on January 20 and 21 is lower. However, the nightly peaks are still present. 

The \ac{MAE}$_{norm}$ of $0.14$ indicates a high \ac{MAE} with respect to the min-max range of the feeder. In the week shown in \Cref{fig:one_feeder_plot}, we observe that the pseudo-measurements show a higher base load compared to the measurements. At the weekend, we observe the lowest pseudo-measurements. In the nights with the lowest temperature below $-5$~\si{\degreeCelsius}, the pseudo-measurements are the highest exceeding $20$~\si{\kW}. In the night with the highest temperatures of the week, it can be observed that the pseudo-measurements are lower compared to the other nights. 

The peak metrics are only evaluated for the consumption due to the constant positive load. The \ac{PMag}$^{C}$ of $10.28$~\si{\kW} is lower compared to the fold average of $11.78$~\si{\kW}. In contrast to the measurements, the pseudo-measurements show lower consumption peaks in the night. The \ac{PTime}$^{C}$ of $6.56$~\si{\hour} and \ac{PShape}$^{C}$ of $0.38$ are higher than the fold average. 

\Cref{fig:one_feeder_plot_f2_summer_period} presents a week for the same feeder $F_2$ but during the summer. We observe that the measurements do not exhibit peak loads during the night and the pseudo-measurements show a lower base load compared to the winter week in \Cref{fig:one_feeder_plot}. 

\section{Discussion}\label{sec:Discussion}
First, we discuss the results with respect to the performance of the presented models. Afterward, we investigate the design of the peak metrics.  

\subsection{Model performance}
In the following, we discuss the performance of the models in comparison with each other and with respect to different folds of the cross-validation. We further argue about possible model reasons for the pseudo-measurement results based on the \ac{XGBoost} model of fold~$1$ as a representative of the examined models. 

\paragraph*{Superiority of nonlinear model}
As already stated in the results section, \ac{XGBoost} and \ac{MLP} outperform \ac{LR}. The models exhibit lower metrics in \Cref{tab:describe_all_observation_metrics,tab:describe_peak_metrics} compared to \ac{LR} except for \ac{PTime}$^{F}$, \ac{PShape}$^{F}$ and some max and min rows. However, the differences of the models regarding \ac{PTime}$^{F}$ and \ac{PShape}$^{F}$ are small and the statistical measures max and min represent only one feeder out of $2{,}323$, respectively. Therefore, we argue that the solution to the problem statement in \Cref{sec:Problem statement} requires a nonlinear model. 

We treat the problem statement as a regression problem on tabular data. It is still unclear whether tree-based models or neural networks are better suited for this \cite{gorishniyRevisitingDeepLearning06}. In our experiments, we cannot identify a superiority of the \ac{XGBoost} model with respect to \ac{MLP} or vice versa. It is noteworthy that the \ac{MLP} with only few parameters consisting of one hidden layer and $20$ neurons performs comparatively to the \ac{XGBoost} model. 


\paragraph*{Challenges in estimating pseudo-measurements}\label{paragraph:Challenges in estimating pseudo-measurements} 
The results show that the distributions of the metrics evaluated on all feeders are right-skewed indicating that the generated pseudo-measurements of some feeders do not fit well to the measurements. 
Therefore, we give four reasons for the deviations. 

First, we observe feeders with assumed data quality issues which are difficult to identify and remedy. An example of this is storage heaters which are registered but not deregistered after decommissioning. 

Second, there are feeders exhibiting measurements which are hard to estimate. We often observe this for measurements exhibiting short but high consumption peaks, for example due to heating systems or commerce and industry. An explanation could be that it is better for models to estimate close to the base load to optimize their quadratic loss function instead of estimating peaks. However, it should be noted that the challenge to estimate peaks is also related to the third and fourth reason given next. 

Third, the features given to the pseudo-measurement model cannot explain all measurements. For example, the feature housing units does not make a statement if the house is a single-family home or a small apartment. This could be the reason for the high \ac{MAE}$_{norm}$ metric of $0.19$ of feeder $F_4$ in the appendix indicating a bad estimation for the base load. Another example is that the only feature for \ac{PV} systems, the installed power, does not cover details about tilt or azimuth angles of the \ac{PV}. This could explain some high values for the metric \ac{PTime}$^{F}$, for example for $F_1$. 

Fourth, the behavior of the customers in a \ac{LV} grid is stochastic. An example is \ac{EV} charging, leading to sudden peaks in the measurements which could be the reason for the peak of feeder $F_1$ at Friday morning in \Cref{fig:one_feeder_plot}. However, the pseudo-measurements represent point estimations which cannot capture the inherent uncertainty of the measurements. 


\paragraph*{High diversity of metadata and measurements}
During the data analysis and experiments, we observe a high diversity both in the feeder metadata as well as in the active power measurements for different seasons. This is visible by the diverse weekly load profiles in \Cref{fig:quantile_plots} and the different weeks of feeder $F_2$ in \Cref{fig:one_feeder_plot} and \Cref{fig:one_feeder_plot_f2_summer_period}. The high diversity is a reason why the performances of the same feeders differ regarding the metrics, for instance the \ac{PMag}$^{C}$ compared to \ac{PMag}$^{F}$ of feeder $F_1$ in \Cref{tab:feeder_metric_details}. 

\paragraph*{Performance with respect to peak metrics}
The \Cref{tab:describe_all_observation_metrics,tab:describe_peak_metrics} illustrate that the \ac{PMag} metrics are worse compared to the \ac{MAE} which is evaluated for all observations. Reasons are stated within this \Cref{paragraph:Challenges in estimating pseudo-measurements} in the paragraph about the challenges in estimating pseudo-measurement. In general, the difficulty of estimating the \ac{PMag} depends heavily on the feeder, the feeder metadata and the time of the year. 

Regarding the PTime, the metric is better for feed-in compared to consumption with respect to mean and median but also regarding the standard deviation. Possible explanations are higher stochastic of consumption compared to feed-in or several consumption peaks on the same day with similar magnitude. This is depicted in \Cref{fig:one_feeder_plot} for feeder $F_1$ on Friday and Saturday where the peak for this day is estimated at the evening instead of the morning. 

The \ac{PShape} metric shows better results for the feed-in compared to the consumption. A reason could be that the shape of a feed-in is mainly induced by \ac{PV} which resembles a bell curve upside down on cloud-free days (compare $F_1$ and $F_3$ in \Cref{fig:quantile_plots}). In contrast, the shapes of the consumption are more diverse. 

\paragraph*{Classification into existing literature}
We cannot directly compare our metric results to other literature, because the papers leveraging grid measurements with feeder metadata, introduced in \Cref{sec:Related Work}, aim to generate load profiles and not complete time-series \cite{adinolfiPseudomeasurementsModelingUsing2014, salazarDataDrivenFramework2020}. Additionally, different temporal resolutions, metrics, grid areas and grid levels make the comparison difficult. In particular, we observe in our pre-experiments that the temporal resolution and the grid level (\ac{MV}/\ac{LV} substation versus \ac{LV} feeder) influences the model accuracy. This is probably due to smoothing effects as stated in \cite{habenReviewLowVoltage2021}. 

\subsection{Peak metric design}\label{subsec:Peak metric design} 
In this section we discuss advantages and constraints of the presented peak metrics. 

\paragraph*{Information gain through peak metrics}
The peak metrics add valuable information to assess the model performance with respect to DSO requirements which often necessitate good peak estimations. \ac{PMag}, \ac{PTime} and \ac{PShape} are suited to assess if a model is appropriate for use cases like preventing overload with control operations. During the experiments, it is important to refine the peak metrics to account for measurements with many zero-crossings and feeders with low capacity utilization. Further refinements of the peak metric design could include adaptations to \ac{PShape} which shows small deviations between feeders and models, especially for the feed-in.

\paragraph*{Combining peak metrics with all-observation metrics}
Analyzing peak metrics in isolation can lead to wrong assumptions when evaluating results. Consider a measurement with a base load and regular peaks. A pseudo-measurement model which estimates a base load close to the peak values of the measurements and not estimating any peaks receives a good peak metric value. However, both the base load and the peaks are estimated badly. Therefore, it is important to also include all-observation metrics in the analysis. 


\paragraph*{Challenge of setting thresholds}
The peak metrics require numerous thresholds for evaluation which must be specified. In particular, the peak metrics are based on daily thresholds which is meaningful because the load follows daily patterns. However, peaks occurring around midnight can lead to high errors of \ac{PMag} and \ac{PTime} even though the estimation is close to the peak with respect to magnitude and timing. 

We also use a threshold of $\pm 10$~\si{\kW} to exclude days from the evaluation which do not exhibit a clear consumption or feed-in peak. This is for example the case for days with low global radiation. The threshold can also be used to evaluate only critical peaks, for example peaks exceeding $80$~\% of the cable capacity. Furthermore, feeders with less than ten days with peaks exceeding $\pm 10$~\si{\kW} are excluded because these feeders can result in high outliers, for example a \ac{PTime} of $0$~\si{\hour} when the timing of the only peak is met exactly. 

Despite the problem of setting thresholds, missing thresholds can also lead to difficult interpretations of a peak metric result. Feeder $F_2$ performs very differently during winter and summer as shown in \Cref{fig:one_feeder_plot,fig:one_feeder_plot_f2_summer_period}. However, this is not reflected in a peak metric averaging all days regardless of the season. 


%



\section{Limitations}\label{sec:Limitations}
In the following, we present the limitations of the dataset, the model and the approach in general.

\paragraph*{Data preparation and dataset} 
The DSO providing the data operates distribution grids mainly in rural areas. Therefore, distribution grids in urban areas are represented less.

Applying filters to the data is necessary, but the concrete thresholds are difficult to define, for instance for feeders with unrealistic high installed equipment power. Furthermore, it is only possible to detect implausible combinations of measurements and feeder meta data if measurements are available for that feeder. However, most feeders are not measured according to the problem statement in \Cref{sec:Problem statement}. This can lead to a bias between training and inference data sets. 

Other challenges regarding the data quality comprise missing information about grid topology changes which affects the feeder metadata or the deregistration of installed equipment after decommissioning. 

We do not apply anomaly removal techniques, because the distinction between anomalies and valid extreme values is ambiguous. However, extreme values are important to estimate peaks and to not deform the test data. Additionally, the aggregation to $15$ minutes smooths short-time anomalies. 

Since the metrics in \Cref{tab:describe_all_observation_metrics,tab:describe_peak_metrics} are calculated after computing the metrics feeder-wise, feeders with long measurement periods could be underrepresented. However, by our approach we can ensure an equal representation of the distribution of metadata at feeders within the metrics. 

\paragraph*{Model}
We do not conduct an extensive hyperparameter optimization for the models of all cross-validation folds. Hence, we do not state if \ac{XGBoost} or \ac{MLP} outperform each other. However, we tested a variety of hyperparameter combinations.

We do not include deep learning approaches. However, we do not expect significant improvements compared to \ac{XGBoost} because tree-based models are state-of-the-art for tabular regression problems \cite{grinsztajnWhyTreebasedModels2022}.

\paragraph{Limitations of the approach}
The presented approach requires well-maintained metadata assigned to \ac{LV} feeders. This excludes the application of the approach in meshed grids where metadata contributes to several feeders. However, \ac{LV} grids are operated predominantly as weakly meshed network with few loops \cite{lourencoPowerDistributionSystem2022}. 

We suspect that optimizing the model regarding all measurement values with quadratic loss functions makes it hard to correctly predict peaks. Nevertheless, several adaptations are possible to mitigate this problem as stated in \Cref{sec:Conclusion}. 

\section{Conclusion}\label{sec:Conclusion}
The lack of active power measurements in the \ac{LV} grid is an urgent problem for \acp{DSO}. In the present paper, we introduce a new approach to generate pseudo-measurements for \ac{LV} feeders which are not equipped with measurement devices. The measurement estimations are based on feeder metadata and adapt to different weather, calendar and timestamp conditions. To the best of our knowledge, there is no other publication applying and analyzing the approach in the given depth.

We extensively evaluate the approach with different models in a $5$-fold cross-validation on real-world data with $2{,}323$ feeders. We show that the approach can produce realistic estimations of measurements. Existing performance outliers are discussed in detail, which occur due to data quality issues, sudden peaks, missing features or stochastics in the data. 

Furthermore, we give insights into the generated pseudo-measurements based on tailored peak metrics. With the help of the introduced peak metrics, we can evaluate the model performance with respect to the magnitude, timing and shape for peaks induced by consumption or feed-in. It is visible that the \ac{MAE} only evaluating peak magnitudes is higher compared to the \ac{MAE} over all observations. 

In the future, the new approach can also be evaluated on other grid levels like substations, other target variables like the current and on different temporal resolutions. Additionally, other datasets with different characteristics and from different countries could be examined. Furthermore, an extensive benchmarking of other approaches to generate estimations for non-measured \ac{LV} feeders could give valuable insights, for example by including the aggregation of smart meter data. Moreover, a comparison between load forecasting and pseudo-measurements as well as using the pseudo-measurements in a state estimation can be examined in the future.

Possible improvements of the presented approach include optimizing the models with respect to the peak metrics or even estimating only peak values. Additionally, probabilistic models could give worst-case load estimates by providing quantiles and thereby estimating the aleatoric uncertainty. The epistemic uncertainty could be reduced by more diverse feeder metadata, for example yearly billing data of households. Finally, post-processing pseudo-measurements could improve the pseudo-measurements with respect to specific metrics. The peak metrics in the present paper provide a basis to evaluate if model adaptations contribute to meet \ac{DSO} requirements. 

Moreover, it is beneficial to generate synthetic data with similar characteristics which are not subject to any legal restrictions and can be published with the results. 


\section*{Acknowledgments}
The authors thank Netze BW GmbH, in particular all team members of the project low voltage prognosis, for the data and infrastructure required for this study. Furthermore, we would like to thank the Helmholtz Association for the support by the Program “Energy System Design” and the Helmholtz Association’s Initiative Helmholtz AI.

\normalsize
\printbibliography

\appendix
\renewcommand\thefigure{\thesection.\arabic{figure}}
\renewcommand\thetable{\thesection.\arabic{table}}
\section{Appendix}
\setcounter{figure}{0}
\setcounter{table}{0}

\begin{figure*}
    \centering
    \includegraphics[width=1\textwidth]{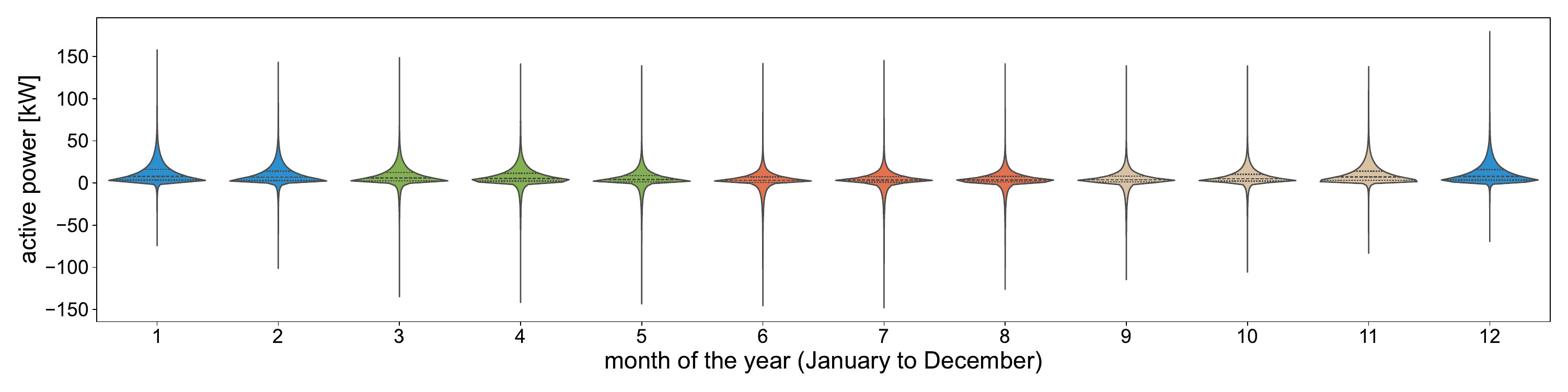}
    \caption{Monthly distributions of measurements from all feeders aggregated to $15$ minute resolution. The colors indicate spring (green), summer (red), autumn (brown) and winter (blue). The dotted lines denote the $25$~\% quantile, median and $75$~\% quantile.}
    \label{fig:violinplot_measurements}
\end{figure*}

\begin{table*}[]
    \centering
    \caption{Overview over the feeder metadata which is used as model feature. Other model features which are not listed are weather and calendar features.}
    \begin{tabular}{llll}
        \toprule
        Group & Type & Meta data features & Unit \\
        \midrule
        Grid connection points & housing & housing units & \# (count)\\
        \midrule
        Installed power of equipment & consumer & \makecell[tl]{storage heaters, heat pumps, electric heaters,\\ \ac{EV} chargers, hot water tanks,\\ inductive power, flow-type heaters,\\ public lighting, other consumers} & \si{\kW}\\
         & battery & batteries & \si{\kW}\\
         & producer & \ac{PV} systems, other producers & \si{\kW}\\
        \midrule
        Energy consumption data & commerce and industry & \makecell[tl]{g0 (general), g1 (workdays), g2 (evening),\\ g3 (continuous), g4 (shop/hairdresser), g5 (bakery),\\ g6 (weekend), l0 (farm)} & \si{\kWh}\\
        \bottomrule
    \end{tabular}
    \label{tab:metadata_features_table}
\end{table*}

\begin{table*}[]
    \centering
    \caption{Hyperparameters used for the models \ac{XGBoost} and \ac{MLP} (without default values).}
    \begin{tabular}{llr|llr}
        \toprule
        Model & Hyperparameter & Value & Model & Hyperparameter & Value\\
         \midrule
        \ac{XGBoost} & n\_estimators & 1500 & \ac{MLP} & hidden\_layer\_sizes & (20)\\
         & eta & 0.05 & & activation & tanh\\
         & subsample & 0.7 & & random\_state & 2\\
         & colsample\_bylevel & 0.5 & & early\_stopping & True\\
         & early\_stopping\_rounds & 30 & & validation\_fraction & 0.125\\
        \bottomrule
    \end{tabular}
    \label{tab:hyperparameters}
\end{table*}

\begin{table*}
    \centering
    \caption{Mean and standard deviation of the $5$ folds from the cross-validation for all metrics and models.}
    \begin{tabular}{l|rrr}
        \toprule    
        Metric & \ac{XGBoost}  & \ac{MLP} & \ac{LR} \\
        \midrule
        \acs{MAE} [\si{\kW}] & 4.73 $\pm$ 0.21 & 4.85 $\pm$ 0.24 & 6.07 $\pm$ 0.29 \\
        \acs{MAE}$_{norm}$ & 0.13 $\pm$ 0.02 & 0.13 $\pm$ 0.02 & 0.17 $\pm$ 0.02 \\
        \acs{RMSE} [\si{\kW}] & 6.19 $\pm$ 0.22 & 6.33 $\pm$ 0.26 & 7.85 $\pm$ 0.49 \\
         \midrule
        $\text{PMag}^{C}$ [\si{\kW}] & 11.68 $\pm$ 0.56 & 11.55 $\pm$ 0.81 & 14.07 $\pm$ 0.72 \\
        $\text{PMag}^{F}$ [\si{\kW}] & 13.24 $\pm$ 0.87 & 12.67 $\pm$ 1.23 & 21.43 $\pm$ 1.66 \\
        $\text{PTime}^{C}$ [\si{\hour}] & 5.09 $\pm$ 0.27 & 5.17 $\pm$ 0.41 & 8.02 $\pm$ 0.19 \\
        $\text{PTime}^{F}$ [\si{\hour}] & 1.17 $\pm$ 0.04 & 1.26 $\pm$ 0.03 & 1.13 $\pm$ 0.02 \\
        $\text{Shape}^{C}$ & 0.35 $\pm$ 0.00 & 0.34 $\pm$ 0.00 & 0.38 $\pm$ 0.00 \\
        $\text{Shape}^{F}$ & 0.29 $\pm$ 0.00 & 0.29 $\pm$ 0.00 & 0.29 $\pm$ 0.00 \\
        \bottomrule
    \end{tabular}
    \label{tab:fold_standard_deviations}
\end{table*}

\begin{table*}
    \caption{Metric overview for the \ac{LV} feeders which are shown in detail in \Cref{fig:one_feeder_plot}, \Cref{fig:quantile_plots} and \Cref{fig:one_feeder_plot_f2_summer_period}. The metric is calculated for the whole measurement period of a feeder. All feeders are part of the test data in the \ac{XGBoost} fold~$1$. The peak metrics are differentiated between consumption (C) and feed-in (F).}
    \centering
    \begin{tabular}{lrrrrrrrrr}
        \toprule
        Feeder & \ac{MAE} & \ac{MAE}$_{norm}$ & RMSE & $\ac{PMag}^{C}$ & $\ac{PMag}^{F}$ & $\ac{PTime}^{C}$ & $\ac{PTime}^{F}$ & $\ac{PShape}^{C}$ & $\ac{PShape}^{F}$ \\
        \midrule
        \makecell[l]{Average\\ XGBoost Fold 1} & 4.57 & 0.10 & 6.22 & 11.78 & 12.58 & 5.11 & 1.18 & 0.35 & 0.30 \\
        \midrule
        $F_1$ (\Cref{fig:one_feeder_plot,fig:quantile_plots}) & 4.62 & 0.04 & 6.75 & 7.14 & 18.46 & 7.38 & 1.10 & 0.32 & 0.29 \\
        $F_2$ (\Cref{fig:one_feeder_plot,fig:one_feeder_plot_f2_summer_period,fig:quantile_plots}) & 4.92 & 0.14 & 6.15 & 10.28 & - & 6.56 & - & 0.38 & - \\
        \midrule
        $F_3$ (\Cref{fig:quantile_plots}) & 6.39 & 0.06 & 8.62 & 5.80 & 11.17 & 1.66 & 1.60 & 0.28 & 0.30 \\
        $F_4$ (\Cref{fig:quantile_plots}) & 6.47 & 0.19 & 7.24 & 6.34 & - & 2.49 & - & 0.32 & - \\
        $F_5$ (\Cref{fig:quantile_plots}) & 12.41 & 0.12 & 14.96 & 30.78 & 19.34 & 3.75 & 3.26 & 0.34 & 0.33 \\
        \bottomrule
    \end{tabular}
    \label{tab:feeder_metric_details}
\end{table*}

\begin{figure*}
    \centering
    \includegraphics[width=1\textwidth]{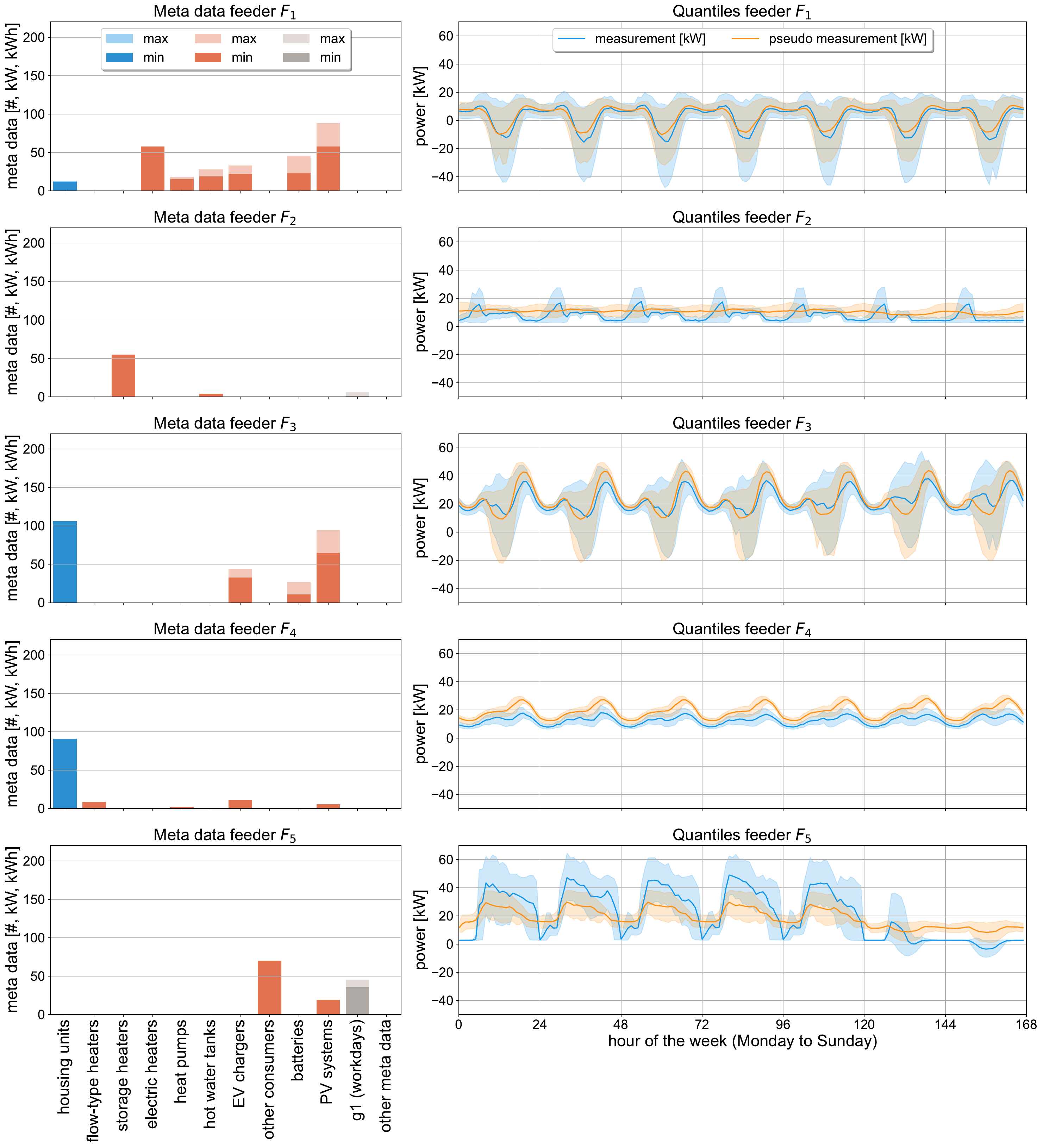}
    \caption{Selected feeders $F_1$ - $F_5$ with feeder metadata, measurements and pseudo-measurements from the test data of \ac{XGBoost} fold~$1$. Metadata with unit count (\#) in blue, unit \si{\kW} in red and unit \si{\kWh} in gray. The label \textit{min} is the lowest value of the metadata and \textit{max} the highest during the evaluation period from the end of 2022 until March 2024. The time series of measurements and pseudo-measurements are condensed to a weekly profile with median, $10$~\% and $90$~\% quantile for visualization.}
    \label{fig:quantile_plots}
\end{figure*}

\begin{figure*}
    \centering
    \includegraphics[width=1\textwidth]{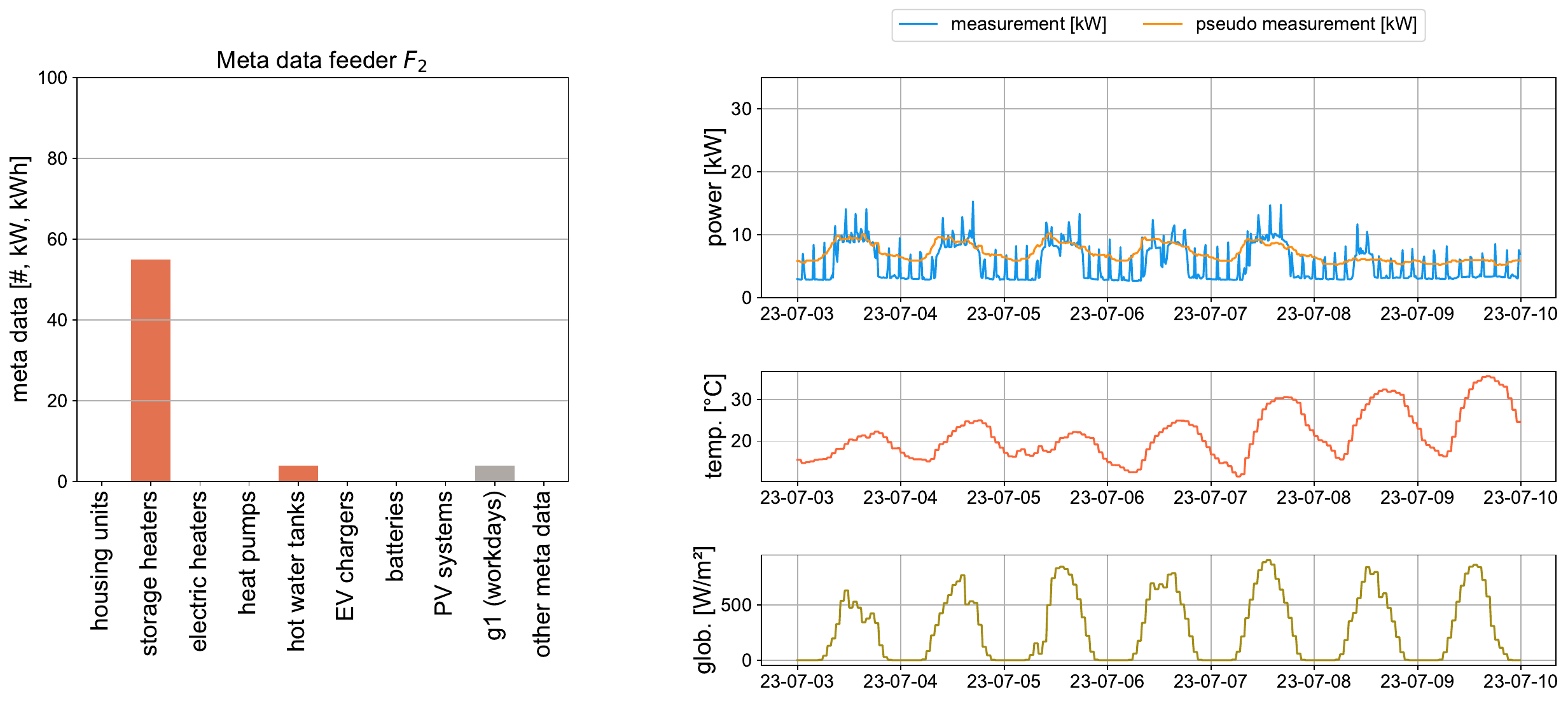}
    \caption{Feeder $F_2$ with the feeder metadata on the left side and the measurement and pseudo-measurement on the right side together with the temperature and global radiation  of one week in July $2023$ (Monday - Sunday). The figure complements the \Cref{fig:one_feeder_plot} where the feeder $F_2$ is shown for a week in winter. The metric values for the feeders are given in \Cref{tab:feeder_metric_details}.}
    \label{fig:one_feeder_plot_f2_summer_period}
\end{figure*}

\end{document}